\documentclass[11pt,a4paper]{article}
\usepackage[hyperref]{emnlp-ijcnlp-2019}
\usepackage{times}
\usepackage{latexsym}

\usepackage{url}

\aclfinalcopy % Uncomment this line for the final submission

\usepackage{array}
\newcolumntype{P}[1]{>{\centering\arraybackslash}p{#1}}

\usepackage{amsmath}
\usepackage{amsfonts}
\usepackage{graphicx}
\usepackage{color}
\usepackage{booktabs}
\usepackage{enumerate}
\usepackage{multirow}

\usepackage{arydshln}
\usepackage[english]{babel}
\usepackage[autostyle, english=american]{csquotes}
\MakeOuterQuote{"}
\usepackage{subcaption}
\usepackage{soul}
\usepackage{mwe}

\title{Automatically Learning Data Augmentation Policies for Dialogue Tasks}

\author{Tong Niu $\;\;\;\;$ Mohit Bansal \\
  UNC Chapel Hill  \\
  {\tt \{tongn, mbansal\}@cs.unc.edu} %\\\And
}

\date{}

\begin{document}
\maketitle

\begin{abstract}
Automatic data augmentation (AutoAugment)~\cite{cubuk2018autoaugment}
searches for optimal perturbation policies via a controller trained using performance rewards of a sampled policy on the target task, hence reducing data-level model bias. While being a powerful algorithm, their work has focused on computer vision tasks, where it is comparatively easy to apply imperceptible perturbations without changing an image's semantic meaning. In our work, we adapt AutoAugment to automatically discover effective perturbation policies for natural language processing (NLP) tasks such as dialogue generation. We start with a pool of atomic operations that apply subtle semantic-preserving perturbations to the source inputs of a dialogue task (e.g., different POS-tag types of stopword dropout, grammatical errors, and
paraphrasing). Next, we allow the controller to learn more complex augmentation policies by searching over the space of the various combinations of these atomic operations. Moreover, we also explore conditioning the controller on the source inputs of the target task, since certain strategies may not apply to inputs that do not contain that strategy's required linguistic features. Empirically, we demonstrate that both our input-agnostic and input-aware controllers discover useful data augmentation policies, and achieve significant improvements over the previous state-of-the-art, including trained on manually-designed policies.

\end{abstract}

\section{Introduction}
Data augmentation aims at teaching invariances to a model so that it generalizes better outside the training set distribution. Recently, there has been substantial progress in Automatic Data Augmentation (\textit{AutoAugment}) for computer vision~\cite{cubuk2018autoaugment}. This algorithm searches for optimal perturbation policies via a controller trained with reinforcement learning, where its reward comes from training the target model with data perturbed by the sampled augmentation policy. Each policy consists of $5$ sub-policies sampled randomly during training, and each sub-policy consists of $2$ operations applied in sequence. These operations are semantic-preserving image processing functions such as \textit{translation} and \textit{rotation}.

\begin{figure}[t]
\centering
\includegraphics[width=0.475\textwidth]{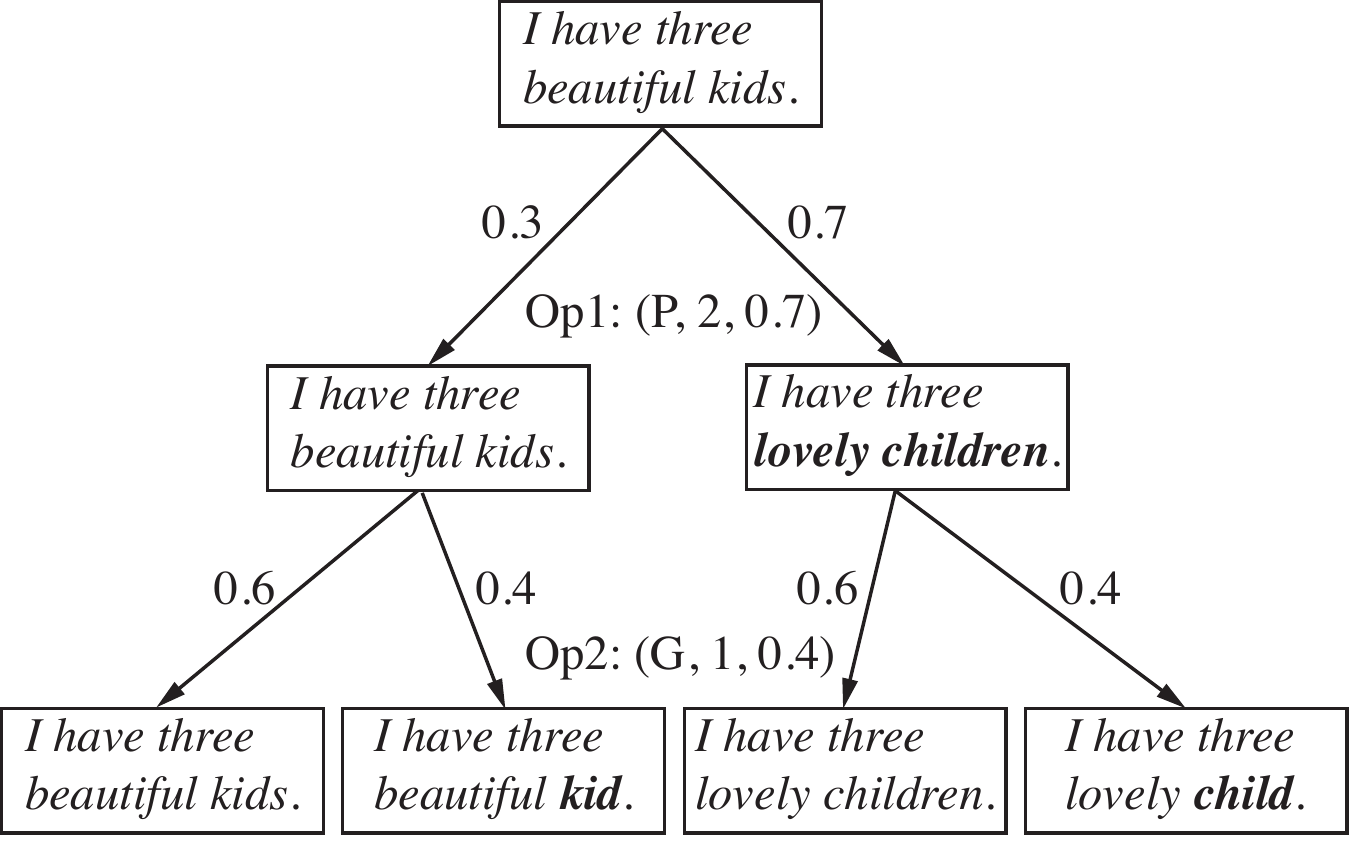}
\vspace{-5pt}
\caption{Example of a sub-policy applied to a source input. The first operation (Paraphrase, $2$, $0.7$) paraphrases the input twice with probability $0.7$; the second operation (Grammar Errors, $1$, $0.4$) inserts $1$ grammar error with probability $0.4$. Thus there are at most $4$ possible outcomes from each sub-policy. 
}
\label{fig:perturbation}
\vspace{-5pt}
\end{figure}

We adapt AutoAugment to NLP tasks, where the operations are subtle, semantic-preserving text perturbations. To collect a pool of such operations, the first challenge we face is that the discrete nature of text makes it less straightforward to come up with semantic-preserving perturbations. We thus employ as a starting point Should-Not-Change strategies (equivalent to \textit{operations} in this paper) proposed by~\newcite{niu2018adversarial} which are shown to improve their dialogue task performance when trained on data perturbed by these strategies. Importantly, we next divide their operations into several smaller ones (such as Grammar Errors divided into Singular/Plural Errors and Verb Inflection Errors) and also add a new operation Stammer (word repetition). This modification provides a much larger space of operation combinations for the model to explore from, so that it could potentially learn more complex and nuanced augmentation policies.
Figure~\ref{fig:perturbation} shows a sub-policy containing two operations. It first paraphrases $2$ tokens with probability $0.7$ and then introduces $1$ grammar error with probability $0.4$.\footnote{Our code and sampled augmented data is publicly available at: \url{https://github.com/WolfNiu/AutoAugDialogue}. The learned policies are presented in Table~\ref{tab:policy}.}

We choose the dialogue generation task based on the Ubuntu Dialogue Corpus~\cite{lowe2015ubuntu} because as opposed to Natural Language Inference and Question Answering tasks, real-world dialogue datasets more naturally afford such perturbation-style human errors (i.e., contain more noise), and thus are inherently compatible with a variety of artificial perturbations.
Empirically, we show that our controller can self-learn policies that achieve state-of-the-art performance on Ubuntu, even compared with very strong baselines such as the best manually-designed policy in~\newcite{niu2018adversarial}. 
We also verify this result through human evaluation to show that our model indeed learns to generate higher-quality responses. We next analyze the best-learned policies and observe that the controller prefers to sample operations which work well on their own as augmentation policies, and then combines them into stronger policy sequences.
Lastly, observing that some operations require the source inputs to have certain linguistic features (e.g., we cannot apply Stopword Dropout to inputs that do not contain stopwords), we also explore a controller that conditions on the source inputs of the target dataset, via a sequence-to-sequence (seq2seq) controller. We show that this input-aware model performs on par with the input-agnostic one (where the controller outputs do not depend on the source inputs), and may need more epochs to expose the model to the many diverse policies it generates.
We also present selected best policies to demonstrate that the seq2seq controller can sometimes successfully attend to the source inputs. 

\section{Related Works}
There has been extensive work that employs data augmentation in both computer vision~\cite{simard2003best,krizhevsky2012imagenet,cirecsan2012multi,wan2013regularization,sato2015apac,devries2017dataset,tran2017bayesian,lemley2017smart} and NLP~\cite{furstenau2009semi,sennrich2015improving,wang2015s,zhang2015character,jia2016data,kim2016sequence,hu2017controllable,xu2017variational,xia2017dual,silfverberg2017data,kafle2017data,hou2018sequence,wang2018switchout}.

Automatic data augmentation is addressed via the AutoAugment algorithm proposed by~\newcite{cubuk2018autoaugment}, which uses a hypernetwork (in our case, a controller) to train the target model, an approach inspired by neural architecture search~\cite{zoph2017learning}.
Previous works have also adopted Generative Adversarial Networks~\cite{goodfellow2014generative} to either directly generate augmented data~\cite{tran2017bayesian,sixt2018rendergan,antoniou2017data,zhu2017data,mun2017generative,perez2017effectiveness}, or generate augmentation strategies~\cite{ratner2017learning}. These approaches produce perturbations through continuous hidden representations. 
Motivated by the fact that our pool of candidate perturbations are discrete in nature, we identify AutoAugment as a more proper base model to use, and adapt it linguistically to the challenging task of generative dialogue tasks. Our work closely follows~\newcite{niu2018adversarial} to obtain a pool of candidate operations.
Although their work also used combinations of operations for data augmentation, their best model is manually designed, training on each operation in sequence, while our model automatically discovers more nuanced and detailed policies that have not only the operation types but also the intensity (the number of changes) and probability. 

\section{Model}
\begin{figure}[t]
\centering
\includegraphics[width=0.4\textwidth]{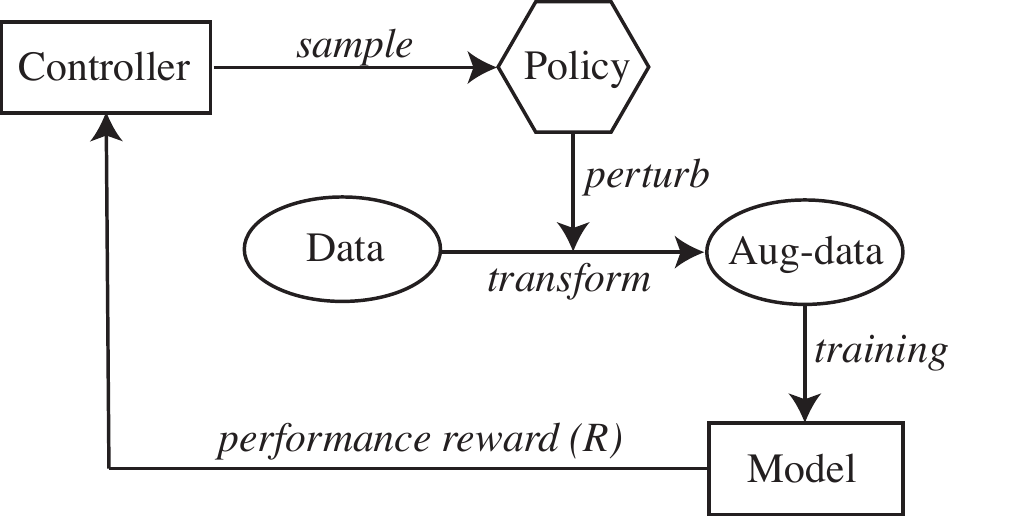}
\vspace{-7pt}
\caption{The controller samples a policy which is used to perturb the training data. After training on the augmented inputs, the model feeds the performance as reward back to the controller.}
\label{fig:controller}
\vspace{-11pt}
\end{figure}

\begin{figure}[t]
\centering
\includegraphics[width=0.34\textwidth]{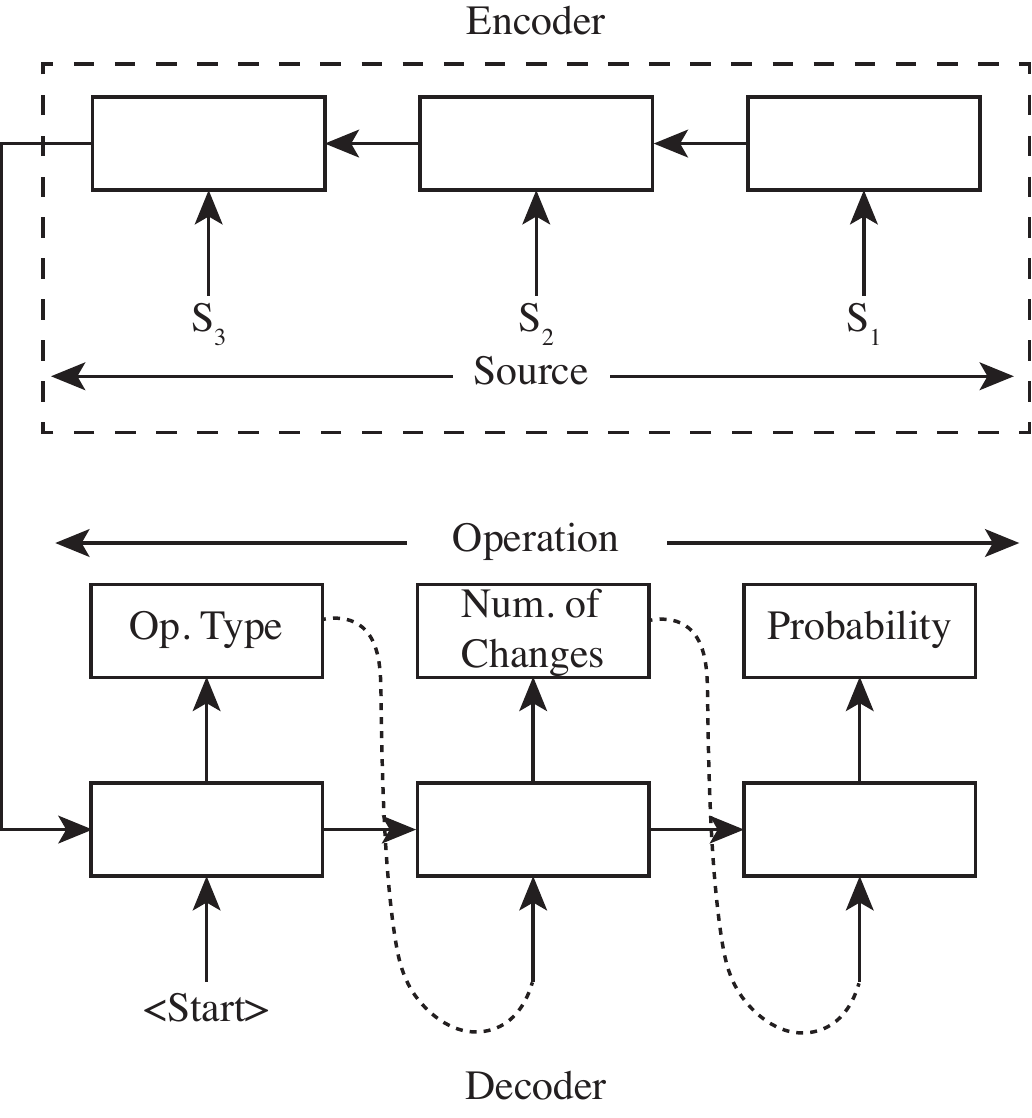}
\vspace{-8pt}
\caption{AutoAugment controller. An input-agnostic controller corresponds to the lower part of the figure. It samples a list of operations in sequence. An input-aware controller additionally has an encoder (upper part) that takes in the source inputs of the data.}
\label{fig:seq2seq}
\vspace{-17pt}
\end{figure}

\noindent\textbf{AutoAugment Architecture}: As shown in Figure~\ref{fig:controller}, our AutoAugment model consists of a controller and a target model~\cite{cubuk2018autoaugment}.
The controller first samples a policy that transforms the original data to augmented data, on which the target model trains. After training, the target model is evaluated to obtain the performance on the validation set. This performance is then fed back to the controller as the reward signal.
Figure~\ref{fig:seq2seq} illustrates the details of input-agnostic and input-aware controllers. The former corresponds to the lower half of the figure. It consists of a single decoder that samples each operation in sequence. An operation is defined by $3$ parameters: \textit{Operation Type}, \textit{Number of Changes} (the maximum number of times allowed to perform the operation, which is a discrete equivalence of the continuous \textit{Magnitude} in~\cite{cubuk2018autoaugment}), and the \textit{Probability} of applying that operation.
The input-aware controller corresponds to the whole figure, i.e., we novelly add in an encoder that takes as inputs the source of the training data, making it a seq2seq model. Since for each source input, there may be a different set of perturbations that are most suitable to it, our input-aware controller aims at providing customized operations for each training example.

\noindent\textbf{Search Space}: 
Following~\newcite{niu2018adversarial}, our pool of operations contains Random Swap, Stopword Dropout, Paraphrase, Grammar Errors, and Stammer, which cover a substantial enough search space of real-world noise in text.\footnote{We did not use their Generative Paraphrasing where \textit{number of changes} (which signifies the intensity of the policy) does not apply to this operation, because this operation can generate only one possible paraphrase for each input.}
To allow the controller to learn more nuanced combinations of operations, we further divide Stopword Dropout into $7$ categories: Noun, Adposition, Pronoun, Adverb, Verb, Determiner, and Other, and divide Grammar Errors into Noun (plural/singular confusion) and Verb (verb inflected/base form confusion). For Stopword Dropout, we chose the first six categories because they are the major universal POS tags in the set closed-class words.
We perform this subdivision also because different categories of an operation can influence the model to different extents. Suppose the original utterance is "\textit{What is the offer?}", dropping the interrogative pronoun "\textit{what}" is more semantic-changing than dropping the determiner "\textit{the}." Our pool thus consists of $12$ operations. For generality, we do not distinguish in advance which operation alone is effective as an augmentation policy on the target dataset, but rather let the controller figure it out. 
Moreover, it is possible that an operation alone is not effective, but when applied in sequence with another operation, they collectively teach the model a new pattern of invariance in the data.

Each policy consists of $4$ sub-policies chosen randomly during training, each sub-policy consists of $2$ operations,\footnote{Having $2$ operations allows more perturbations than exerted by a single strategy: in reality a sentence may contain several types of noise.} and each operation has $3$ hyperparameters. We discretize the Probability of applying an operation into $10$ uniformly-spaced values starting from $0.1$ and let the number of replaces range from $1$ to $4$. Thus, the search space size is $(12 \times 4 \times 10)^{(2 \times 4)} = 2.82 \times 10^{21}$. “2 operations” allows more perturbations than exerted by a single strategy: in reality a sentence may contain several types of noise.

\noindent\textbf{Search Algorithm}: We use REINFORCE, a policy gradient method~\cite{williams1992simple,Sutton00policygradient} 
to train the controller.
At each step, the decoder samples a token and feeds it into the next step. Since each policy consists of $4$ sub-policies, each sub-policy contains $2$ operations, and each operation is defined by $3$ tokens (Operation Type, Number of Changes, Probability), the total number of steps for each policy is $4 \times 2 \times 3 = 24$. Sampling multiple sub-policies to form one policy provides us with a less biased estimation of the controller performance. We sample these $4$ sub-policies at once rather than sample the controller $4$ times to reduce repetition -- the controller needs to keep track of what policies it has already sampled. To obtain the reward for the controller, we train the target model with the augmented data and obtain its validation set performance. We calculate a weighted average of two F1s (Activity and Entity) as the reward $R$ since both are important complementary measures of an informative response, as discussed in~\newcite{serban2017multiresolution}.\footnote{We use the weighted average F1 = Activity F1 + $5.94$ $/$ $3.52$ $*$ Entity F1, where the weights are determined by the baseline results, for balance between the two F1s.} 
We use the reinforcement loss following~\newcite{ranzato2015sequence} and an exponential moving average baseline to reduce training variance. 
At the end of the search, we use the best policy to train the target model from scratch and evaluate on the test set.

\section{Experimental Setup}
\label{sect:Experimental Setup}

\noindent\textbf{Dataset and Model}:
We investigate Variational Hierarchical Encoder-Decoder (VHRED)~\cite{serban2017hierarchical} on the troubleshooting Ubuntu Dialogue task~\cite{lowe2015ubuntu}.\footnote{\url{http://www.iulianserban.com/Files/UbuntuDialogueCorpus.zip}} For this short paper, we focus on this dataset because it has a well-established F1 metric among dialogue tasks.\footnote{\newcite{serban2017multiresolution} found that F1 is “particularly suited for the goal-oriented Ubuntu
Dialogue Corpus” based on manual inspection of the extracted activities and entities.} 
In future work, we plan to apply the idea to other datasets and NLP tasks since the proposed model is not specific to the Ubuntu dataset. We hypothesize that most web/online datasets that have regular human errors/noise (e.g. Twitter Dialogue Dataset~\cite{serban2017hierarchical} and multiple movie corpora~\cite{serban2016building}) would be suitable for our framework.

\noindent\textbf{Training Details}:
We employ code\footnote{\url{https://github.com/WolfNiu/AdversarialDialogue}} from~\newcite{niu2018adversarial} to reproduce the VHRED baseline results and follow methods described in~\newcite{zoph2017learning} to train the controller. For details, please refer to their Appendix A.2 (Controller Architecture) and A.3 (Training of the Controller).

We adopt the following method to speed up controller training and hence facilitate scaling.\footnote{It took us $3$ days with $6$ P100 GPUs to get our reported results for each of the input-aware and input-agnostic models. The VHRED model took $1.5$ days to pre-train.} We let the target model resume from the converged baseline checkpoint and train on the perturbed data for $1$ epoch. During testing, we use the policy that achieves the highest weighted average F1 score to train the final model for $1$ epoch. For the All-operations model (which corresponds to the All-Should-Not-Change model in~\newcite{niu2018adversarial}), we train on each operation (without subdivisions for Stopword Dropout and Grammar Errors) for $1$ epoch in sequence.\footnote{We do not follow~\newcite{cubuk2018autoaugment} to train on a small portion of the data for many epochs because we empirically found that this approach decreases our task performance.}

\noindent\textbf{Automatic Evaluation}:
We follow~\newcite{serban2017multiresolution} and evaluate on F1s for both activities (technical verbs) and entities (technical nouns).

\noindent\textbf{Human Evaluation}:
We conducted human studies on MTurk. We compared each of the input-agnostic/aware models with the VHRED baseline and All-operations from~\newcite{niu2018adversarial}, where we followed the same setting. 
Each study contains $100$ samples randomly chosen from the test set. The utterances were randomly shuffled and we only allowed US-located human evaluators with approval rate $> 98\%$, and at least $10,000$ approved HITs. 
More details are in the appendix.
\begin{table}[t]
\small
\centering
\begin{tabular}{ccc}
\toprule
& Activity F1 & Entity F1 \\
\midrule
LSTM & 1.18 & 0.87 \\
HRED & 4.34 & 2.22 \\
VHRED & 4.63 & 2.53 \\
VHRED (w/ attn.) & 5.94 & 3.52 \\
\hdashline
All-operations & 6.53 & 3.79 \\
Input-aware & \textbf{7.04} & 3.90 \\
Input-agnostic & 7.02 & \textbf{4.00} \\
\bottomrule
\end{tabular}
\vspace{-5pt}
\caption{Activity, Entity F1 results reported by previous work (rows $1$-$4$ from~\newcite{serban2017multiresolution}), and the All-operations and AutoAugment models.
}
\label{tab:f1-result}
\vspace{0pt}
\end{table}

\begin{table}[t]
\small
\centering
\begin{tabular}{ccccc}
\toprule
	& W & T & L & W - L \\
\midrule
Input-agnostic vs. baseline  & 48 & 23 & 29 & 19 \\
Input-aware vs. baseline & 45 & 27 & 28 & 17 \\
Input-agnostic vs. All-ops & 43 & 27 & 30 & 13 \\
Input-aware vs. All-ops & 50 & 13 & 37 & 13 \\
\bottomrule
\end{tabular}
\vspace{-5pt}
\caption{Human evaluation results on comparisons among the baseline, All-operations, and the two AutoAugment models. \textbf{W}: Win, \textbf{T}: Tie, \textbf{L}: Loss.}
\label{tab:human-eval}
\vspace{-7pt}
\end{table}

\begin{table*}[t]
\small
  \centering
    \begin{tabular}{cP{0.095\textwidth}p{0.66\textwidth}}
    \toprule
    Perturbation Method & Policy & Source Inputs \\
    \midrule
	Original Context & - & fresh install of crack of the day : gdm login $\rightarrow$ " can't access ACPI bla bla bla " \_\_eou\_\_ you don't want to be me ... \_\_eou\_\_ ah , it happened to you too ? \\
    \hdashline
    All-operations & (R\phantom{v}, 4, 1.0) & fresh install \textbf{crack of} of the day : gdm login $\rightarrow$ " can't \textbf{ACPI access} bla bla bla " \_\_eou\_\_ you don't want \textbf{be to} me ... \_\_eou\_\_ ah , it happened \textbf{you to} too ? \\
    Input-agnostic & (D$_v$, 3, 0.2) (R\phantom{v}, 1, 0.5) & fresh install of crack of the day : \textbf{login gdm} $\rightarrow$ " can't access ACPI bla bla bla " \_\_eou\_\_ you \textbf{\st{don't}} want to \textbf{\st{be}} me ... \_\_eou\_\_ ah , it happened to you too ? \\
    Input-aware & (S\phantom{v}, 1, 0.8) (D$_v$, 2, 0.5) & fresh install of crack of the day : gdm login $\rightarrow$ " can't access ACPI bla bla \textbf{bla} bla " \_\_eou\_\_ you don't want to be me ... \_\_eou\_\_ ah , it happened to you too ? \\
    \bottomrule
    \end{tabular}
    \caption{Comparisons of perturbed outputs among the three Augmentation models. Note that only the sampled/applied sub-policy is shown for each model. The dashed line separates the original and augmented source inputs. Perturbed tokens are boldfaced.}
  \label{tab:example}
\end{table*}

\section{Results and Analysis}
\label{sect:Results}

\noindent\textbf{Automatic Results}:
\label{subsect:Automatic Results}
Table~\ref{tab:f1-result} shows that all data-augmentation approaches (last $3$ rows) improve statistically significantly ($p < 0.01$)
\footnote{Statistical significance is computed via the bootstrap test~\cite{noreen1989computer} over 
100K samples.} over the strongest baseline VHRED (w/ attention). Moreover, our input-agnostic AutoAugment is 
statistically significantly ($p < 0.01$) better (on Activity and Entity F1) than the strong manual-policy All-operations model, while the input-aware model is stat. signif. ($p < 0.01$) better on Activity F1.\footnote{Note that F1s are overall low (reported baselines in Table~\ref{tab:f1-result} are from~\newcite{serban2017multiresolution}) because Ubuntu is semi-task-oriented; each context can have many valid responses containing different activities/entities from the ground-truth.}

\noindent\textbf{Human Evaluation}:
\label{subsect:Human Evaluation}
In Table~\ref{tab:human-eval}, both AutoAugment models obtained significantly more net wins (last column) than the VHRED-attn baseline. They both outperform even the strong manual-policy All-operations model.

\begin{table}[t]
\small
  \centering
  \resizebox{\linewidth}{!}{
    \begin{tabular}{cccc}
    \toprule
     Sub-policy1 & Sub-policy2 & Sub-policy3 & Sub-policy4 \\
    \midrule
     P, 1, 0.5 & D$_v$, 3, 0.2 & R, 3, 0.9 & D$_p$, 2, 0.3  \\
           D$_{adv}$, 4, 0.4 & R, 1, 0.5 & D$_{adp}$, 1, 0.5 & D$_{adp}$, 2, 0.1  \\
    \midrule
     D$_n$, 1, 0.8  & D$_o$, 3, 1.0      &  P, 4, 0.4     &  G$_n$, 3, 0.3      \\
           G$_v$, 1, 0.9     & D$_o$, 3, 0.1      &  S, 3, 0.4      &  R, 1, 0.2      \\
    \midrule
     D$_v$, 2, 0.5     &  D$_v$, 2, 0.7     &   S, 3, 0.5    &  P, 1, 1.0    \\
          R, 2, 0.2      &  G$_v$, 1, 0.9     & D$_o$, 1, 0.5      &  G$_n$, 2, 0.6       \\
    \bottomrule
    \end{tabular}%
  }
  \vspace{-5pt}
  \caption{Top $3$ policies on the validation set and their test performances. 
\textbf{Operations}: R=Random Swap, D=Stopword Dropout, P=Paraphrase, G=Grammar Errors, S=Stammer. \textbf{Universal tags}: n=noun, v=verb, p=pronoun, adv=adverb, adp=adposition.}
  \label{tab:policy}
\vspace{-5pt} 
\end{table}

\noindent\textbf{Policy Learned by the Input-Agnostic Controller}:
We present $3$ best learned policies from the Ubuntu val set (Table~\ref{tab:policy}).
Although there is a probability of $7 / 12 = 58.3\%$ to sample one of the Stopword Dropout operations from our pool, all $3$ learned policies show much more diversity on the operations they choose. 
This is also the case for the other two hyperparameters: \textit{Number of Changes} varies from $1$ to $4$, and \textit{Probability} varies from $0.1$ to $0.9$, which basically extend their whole search range.
Moreover, all best policies include Random Swap, which agrees with the results in~\newcite{niu2018adversarial}.

\vspace{2pt}
\noindent\textbf{Example Analysis of Perturbation Procedure in Generated Responses}:
We also present a selected example of perturbed source inputs from the three Augmentation models with their respective best policies in Table~\ref{tab:example}. First of all, the All-operations model is forced to use an operation (in this case Random Swap) with a fixed number of changes and a probability of $1.0$, leading to less variation in the source inputs. On the other hand, our Input-agnostic AutoAugment model samples 3 Verb Dropout's followed by Random Swap. Note that although the number of changes for the dropout is $3$, there are only $2$ verb stopwords in the utterance. Thus, it has to resort back to modifying only $2$ tokens. The Input-aware model samples Stammer followed by 2 Verb Dropout's. Interestingly, it inserts an extra "\textit{bla}" around other "\textit{bla}'s" in the utterance. It also did not sample a policy that drops more than $2$ verb stopwords (this operation is not applied due to its Probability parameter). These two observations indicate that the model can sometimes successfully attend to the source inputs to provide customized policies.

\section{Conclusion and Future Work}
We adapt AutoAugment to dialogue and extend its controller to a conditioned model. We show via automatic and human evaluations that our AutoAugment models learn useful augmentation policies which lead to state-of-the-art results on the Ubuntu task. Motivated by the promising success of our model in this short paper, we will apply it to other diverse NLP tasks in future work.

\section*{Acknowledgments}
\vspace{-3pt}
We thank the reviewers for their helpful comments and discussions.
This work was supported by DARPA \#YFA17-D17AP00022, ONR Grant \#N00014-18-1-2871, and awards from Google, Facebook, Salesforce, and Adobe (plus Amazon and Google GPU cloud credits). The views are those of the authors and not of the funding agency.

\bibliography{main.bib}
\bibliographystyle{acl_natbib}

\appendix

\section{Experimental Setup}
\label{sect:Experimental Setup Appendix}
\noindent\textbf{Dataset and Model}:
We investigate Variational Hierarchical Encoder-Decoder (VHRED)~\cite{serban2017hierarchical}, a state-of-the-art dialogue model on the Ubuntu Dialogue Corpus~\cite{lowe2015ubuntu}. Ubuntu is a troubleshooting dialogue dataset containing $1$ million 2-person, multi-turn dialogues extracted from Ubuntu chat. The chat channels are used by customers to provide and receive technical support. We focus on the task of generating fluent, relevant, and goal-oriented responses. Following~\newcite{niu2018adversarial}, we apply additive attention mechanism~\cite{bahdanau2014neural} to the source sequence while keeping the remaining VHRED architecture unchanged.

\noindent\textbf{Automatic Evaluation}:
We follow~\newcite{serban2017multiresolution} and evaluate the model on F1's for both activities (technical verbs, e.g., "boot", "delete") and entities (technical nouns, e.g., "root", "disk") computed by mapping the ground-truth and model responses to their corresponding activity-entity representations.

\noindent\textbf{Human Evaluation:}
We conducted human studies on MTurk to evaluate the dialogue quality of generated responses from the investigated models. We compared each of the input-agnostic/aware models with All-operations and the baseline (i.e., $4$ experiments in total). Studies involving the baseline are for sanity checks.

\end{document}